%% file: ICLR.tex
\documentclass{article} 
\usepackage{iclr2021_conference,times}

\input{math_commands.tex}

\usepackage{hyperref}
\usepackage{url}

\usepackage{enumitem}
\usepackage{graphicx}
\graphicspath{{figures/}} 
\graphicspath{{tables/}} 
\graphicspath{{images/}} 

\title{Visual Crowd Analysis: Open Research Problems}

\author{Muhammad Asif Khan\thanks{Corresponding author: Alternate email: asifk@ieee.org}, Hamid Menouar \\
Qatar Mobility Innovations Center \thanks{QMIC is a trade name of Qatar University-QSTP-B, www.qmic.com}\\
Qatar University\\
Doha, Qatar \\
\texttt{\{muhammada, hamidm\}@qmic.com} \\
\And
Ridha Hamila \\
Department of Electrical Engineering \\
Qatar University \\
Doha, Qatar. \\
\texttt{hamila@qu.edu.qa} \\
}

\iclrfinalcopy 

\usepackage[nobiblatex]{xurl}
\usepackage {hyperref}
\hypersetup{hidelinks, colorlinks=true, citecolor=blue}
\usepackage{subcaption}
\usepackage{xcolor}
\usepackage{booktabs}

\definecolor{color1}{HTML}{FFFFFF} 
\definecolor{color2}{HTML}{F9F1A4}
\definecolor{color3}{HTML}{D70040}

\usepackage{soul}
\newcommand{\rev}[1]{\textcolor{black}{#1}} 
\newcommand{\khan}[1]{\textcolor{black}{#1}} 

\begin{document}

\maketitle

\input{abstract2}

\input{main}

\subsubsection*{Acknowledgments}
This publication was made possible by the PDRA award
PDRA7-0606-21012 from the Qatar National Research Fund (a
member of The Qatar Foundation). The statements made herein
are solely the responsibility of the authors.

\bibliography{biblio}
\bibliographystyle{iclr2021_conference}


\end{document}

%% file: math_commands.tex
\usepackage{amsmath,amsfonts,bm}

\def\eqref#1{equation~\ref{#1}}

\def\1{\bm{1}}

\DeclareMathAlphabet{\mathsfit}{\encodingdefault}{\sfdefault}{m}{sl}
\SetMathAlphabet{\mathsfit}{bold}{\encodingdefault}{\sfdefault}{bx}{n}



%% file: abstract2.tex
Over the last decade, there has been a remarkable surge in interest in automated crowd monitoring within the computer vision community. Modern deep-learning approaches have made it possible to develop fully-automated vision-based crowd-monitoring applications. However, despite the magnitude of the issue at hand, the significant technological advancements, and the consistent interest of the research community, there are still numerous challenges that need to be overcome. In this article, we delve into six major areas of visual crowd analysis, emphasizing the key developments in each of these areas. We outline the crucial unresolved issues that must be tackled in future works, in order to ensure that the field of automated crowd monitoring continues to progress and thrive. Several surveys related to this topic have been conducted in the past. Nonetheless, this article thoroughly examines and presents a more intuitive categorization of works, while also depicting the latest breakthroughs within the field, incorporating more recent studies carried out within the last few years in a concise manner. By carefully choosing prominent works with significant contributions in terms of novelty or performance gains, this paper presents a more comprehensive exposition of advancements in the current state-of-the-art.

%% file: main.tex
\section{Introduction}\label{sec:intro}
The increasing population in urban areas often leads to crowded situations in densely populated areas which pose several challenges and threats to public safety. To ensure the safety of the people, crowd management strategies require efficient crowd analysis. While manual analysis of the crowd is a tedious task, automatic crowd analysis is desired in many situations.
\par
Automatic crowd monitoring using visual analysis is a hot topic in computer vision research \cite{khan2022revisiting} with many interesting applications in city surveillance, social distancing, transportation, sports, wildlife monitoring, etc. \cite{reviewer21, reviewer22}. Over the last decade, advances in deep learning have brought new possibilities to achieve state-of-the-art performances in various visual crowd analysis problems such as crowd counting, object detection, activity recognition, anomaly detection, motion analysis, etc.
\khan{Although numerous research efforts have achieved remarkable performance in visual crowd analysis, there remain several challenges and problems yet to address. The reason for many of the unsolved problems lies in the underlying complexity and challenges related to crowd scenes as compared to other computer vision tasks e.g., severe occlusions, clutters, scale variations, unpredictable motion patterns, complex crown behaviors, the unknown context of crowd activities, etc. Thus, crowd analysis is often considered more challenging than other computer vision tasks. The complex nature of the problem impacts the development of the crowd visual analysis system and requires more sophisticated algorithms and models, collection of diverse and large-scale datasets, hardware resources for real-time performance, and system-level integration of such algorithms with cameras, sensors, and data storage systems. Algorithmic innovation and data availability are particularly of paramount importance for computer vision researchers.
}\par
\begin{figure*}
    \centering
    \includegraphics[width=0.99\textwidth]{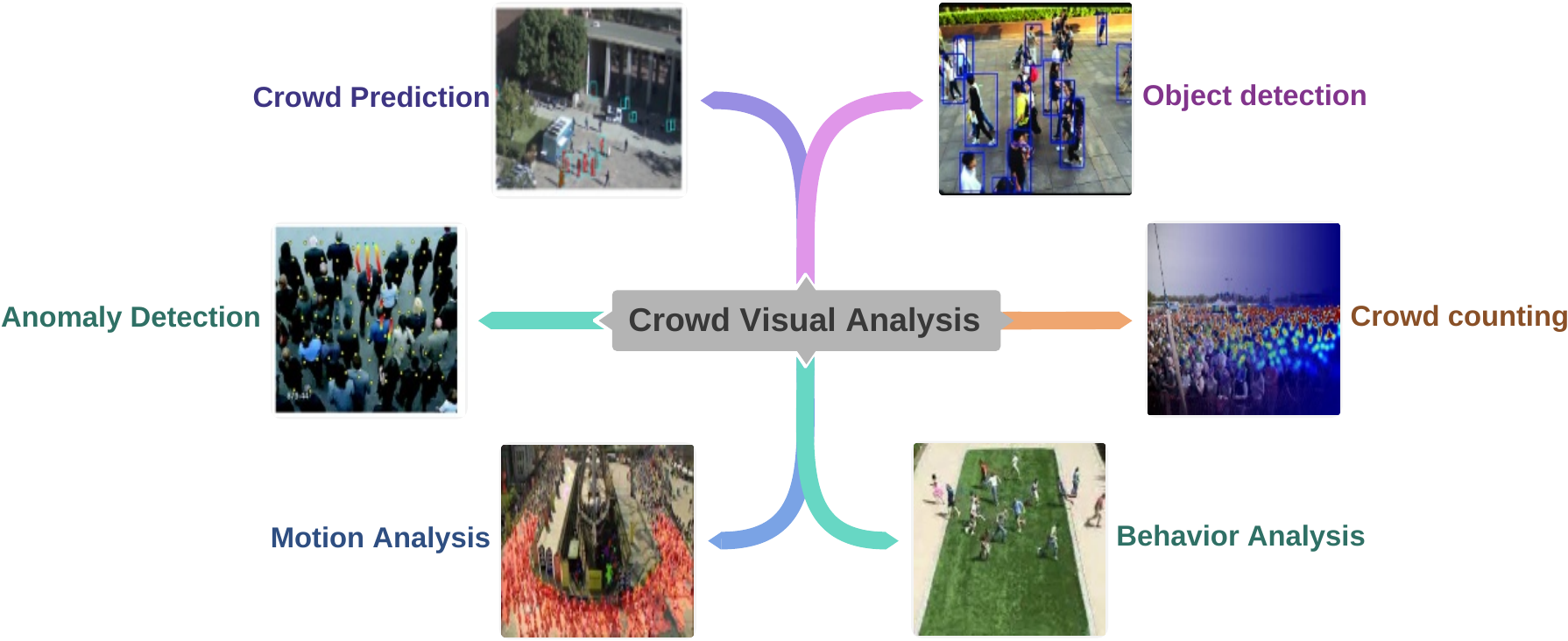}
    \caption{Six areas or applications of crowd visual analysis.}
    \label{fig1}
\end{figure*}

\khan{
To understand these challenges, we thoroughly investigated the literature on crowd analysis using computer vision and mostly using deep learning. We also studied different traditional approaches, to understand how the innovations progressed over time and how they created an impact. We greatly focused on research works published in major scientific venues and chose those works that have significantly contributed to the body of knowledge in terms of identifying a real problem, innovation in methodology, or claiming significant performance gains against previous methods. We found existing surveys categorizing works in different ways, however, when studying the common trends and distinct approaches in these works, we found a more intuitive and meaningful categorization of crowd analysis. Our approach is to categorize these works into different types of analysis tasks where each task is completely different and requires different methods to accomplish the task objective.
As illustrated in Fig. \ref{fig1}, these works on visual crowd analysis are categorized into six major areas i.e., crowd counting, object detection and tracking, motion analysis, behavior recognition, anomaly detection, and crowd prediction.}
\par

\khan{
Crowd counting is to estimate the density of people or the total count in a particular geographical area.
Objection detection aims to detect and localize particular objects of interest in the crowd e.g., detecting women only, detecting people holding banners or sticks, etc.
Motion analysis refers to the collective mobility state of the crowd e.g., if the crowd is stationary/moving along with other motion statistics such as direction, speed, flux analysis, etc.
Behavior analysis determines the collective attribute of the crowd focusing on the activities performed by crowd members to extract contextual behavior information e.g. if the crowd is calm, violent, etc.
Anomaly detection focuses on finding unusual and abnormal events and activities at both individual and group levels.
Crowd prediction refers to the prediction of proactive accumulation (assembly) or influx and efflux of people in/from a particular region that can lead to a crowd.
}
Generally, there is a logical sequence in the implementation of these applications. For instance, the first information which might interest a user or agency monitoring a crowd is the estimated crowd density, followed by other aspects of crowd members such as age, gender, etc., and any detecting objects such as banners, posters, and sticks held by the members. The next information one might be interested in is continuously following up on the crowd movement to know whether the crowd is stationary or moving in a specific direction. The aforementioned aspects are covered in the first three areas of crowd analysis. Then, a sophisticated system may provide a more detailed analysis such as crowd behavior (mood, specific activities performed by crowd members), and detecting specific abnormal events or objects. Lastly, the prediction about crowd formation or dispersion in a specific region or area at a specific time can be a piece of very useful information that one would desire to obtain.
\par
\khan{
While traditionally, many semi-automated computer vision methods have been proposed \cite{Davies1995, cv_crowd2010, Silveira_2010}, the recent advancements in modern deep learning have revolutionized the development of fully automated vision-based crowd-monitoring applications. By leveraging the power of deep neural networks the accuracy, efficiency, and overall performance of such applications have been significantly improved. Deep learning methods automatically learn and extract meaningful patterns and features from large-scale crowd visual data such as images and videos, thus enabling better crowd analysis. Also, the use of transfer learning allows deploying models trained on one dataset in a different scenario after fine-tuning, speeding up the training process.
}

\rev{
\subsection{Similar and Related Studies}
Several studies have been conducted in the past which typically focus on individual areas of crowd analysis. For example, \cite{Sindagi_2018, Cenggoro_2019, Naveed_2020, Gao_2020, Luo_2020, Gouiaa_2021, Fan_2022, khan2022revisiting} covers crowd counting research and mainly discusses the advancements in model architectures, benchmarking, and datasets. \cite{Hu2004ASO, Luca2020ASO, Kumar2021CrowdBM} focus on crowd motion analysis, discussing crowd motion predictions, flow classification, and  behavior analysis using motion patterns. \cite{Joshi2021ACB, Modi2022ASO, Sharif2022DeepCA} studies anomaly detection with focus on methods, datasets and comparisons of results. Unlike the aforementioned studies, few studies also cover the multiple aspects of crowd analysis. For instance, \cite{Li2015CrowdedSA} covers motion analysis, behavior recognition, and anomaly detection but do not cover counting, density estimation, object detection, crowd prediction etc. \cite{HY2017CrowdBA} covers density estimation, motion detection, and behavior recognition, but do not cover object detection, and anomaly detection. Similarly, \cite{Zhang2018PhysicsIM} covers only on physics-inspired methods for crowd analysis and focuses on motion analysis in videos.} Table \ref{tab-surveys} presents a list of recent survey articles on the topic.
\par
\rev{
Although the studies discussed earlier provide a thorough review of the literature with useful taxonomy of works, our study presents a more comprehensive review consisting of the six intuitive areas of crowd analysis. Our survey is unique from the previous surveys as it does not provide a thorough review covering each and every related work published, but focuses on the most prominent works in each application area. Thus this review is more concise, allowing even non-experts and novice researchers in the area of crowd analysis to quickly understand the state-of-the-art in research on the respective topic. Due to the fast-paced research in computer vision, several related works published a few years before may not serve the purpose and yet another new study with fresh perspectives and insights is highly commendable. 
}

\input{tables/tab_surveys2}

\rev{
\subsection{Contribution and Paper Organization}
Although several studies exist as stated in the previous section, discussing individual areas of crowd-monitoring applications, we aim to provide a concise study on the topic targeted to a broader audience from various disciplines. We intentionally omitted more technical details and focused more on the crux of the problem for a smooth flow for the reader. For instance, there have been more than a hundred crowd-counting models published in the literature, we chose only those with significant contributions in terms of novelty in model architecture, performance gains, or other aspects of design and evaluation. Similarly, we carefully chose original research works with major contributions (despite if the work is overlooked previously) in each application domain and included them in our survey to summarise the SOTA in each of the six areas of crowd analysis.}
\par
\rev{
This article discusses the six intuitive areas of crowd visual analysis with a concise description of each, provides a brief overview of the state-of-the-art methods, and highlights the open problems in each domain. Each subsequent section describes one of the six areas. The last section draws the conclusion and presents future insights and research directions to advance the SOTA in the respective area.
}

\section{Counting and Density Estimation}
One of the first aspects of crowd monitoring is to estimate the headcount (a scalar value for the whole image) or density across different parts of the scene. As people in the crowd are usually clumped together into groups, density estimates provide more information than just the total count. Headcount or density estimation can provide good situational awareness for the monitoring entities like law enforcement agencies and event managers.

\subsection{Methods and State-of-the-art}
\begin{figure}[!h]
    \centering
    \includegraphics[width=0.8\textwidth]{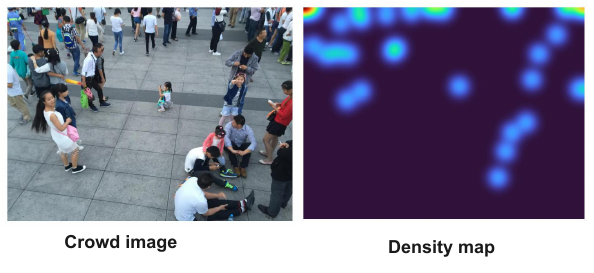}
    \caption{Crowd counting using density estimation. The predicted density map shows crowd density across the image scene. The sum of all pixel values of the density map equals the predicted count in the image.}
    \label{fig:counting}
\end{figure}

The de-facto method for counting people in an image or video frame is \textit{density estimation}. A convolution neural network (CNN) based model is trained to estimate the crowd density.
In the density estimation method, each head is detected using a Gaussian blob around the center of the head. This is a pixel-level regression problem and the commonly used Euclidean loss function (or mean squared error MSE) is used to train the model.
Fig. \ref{fig:counting} presents the ground truth density map for a crowd image used by the CNN model. 

The first CNN-based crowd density estimation model was CrowdCNN \cite{CrowdCNN_CVPR2015}. 
\rev{The CrowdCNN is a single-column CNN architecture that outperforms traditional non-CNN methods however, still lacks the capability to adapt to the scale variations in head sizes.} Scale variation can arise from several reasons such as distance from the camera, camera perspective effects, image resolution, etc. \rev{To overcome scale variations, multi-column CNN networks were proposed \cite{MCNN_CVPR2016, CMTL_AVSS2017}. Multi-column models can capture scale variations to some extent (i.e., more columns are needed for images with larger scale variations which may increase the model size significantly.). However, more efficient architectures were proposed} to replace multi-column architectures e.g., encoder-decoder networks \cite{TEDnet_CVPR2019, SASNet_2021, MobileCount_PRCV2019}, networks with multi-scale modules \cite{MSCNN_ICIP2017, SGANet_IEEEITS2022} etc.
\rev{Encoder-decoder models allow hierarchical feature extraction and aggregation at multiple stages in the encoder and decoder modules, respectively. These models are good at preserving the spatial resolution of the predicted density maps using downsampling and upsampling operations, however, take longer times to train and converge. Recently, non-CNN models using vision transformers are also proposed \cite{TransCrowd_2022, CCTrans_2021}. Transformer-based models apply self-attention mechanisms to model global context and capture long-range dependencies. Their disadvantage is that these models require more computational resources compared to CNNs due to the larger number of parameters and self-attention operations.}

\begin{figure}
    \centering
    \includegraphics[width=0.8\textwidth]{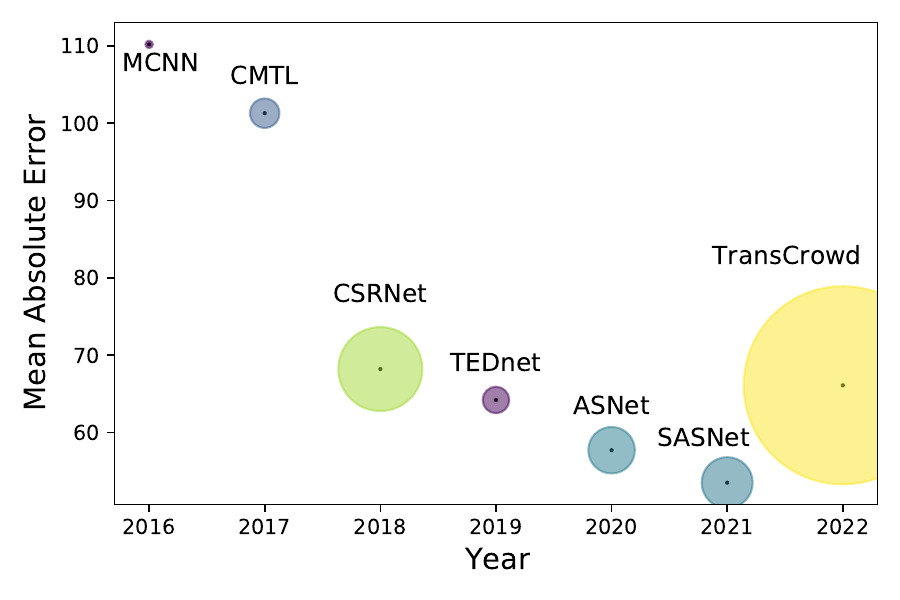}
    \caption{Performance of well-known crowd counting models over years.} \label{fig:mae_over_years}
\end{figure}

Crowd counting is the relatively most rigorously investigated area after object detection with the availability of a large number of datasets (\rev{Mall \cite{Mall_dataset2012}, ShanghaiTech \cite{MCNN_CVPR2016}, UCF-QNRF \cite{CompositionLoss_2018}, JHU-Crowd \cite{jhucrowd_dataset2020}, NWPU-Crowd \cite{NWPUCrowd_dataset2021}, DroneRGBT \cite{DroneRGBT_dataset} etc.}) and a large variety of models with reasonably good performance. \rev{The performance gains in the counting accuracy have been high initially when single-column shallow models have been replaced by multi-column CNN, and then pyramids structure models using transfer learning and multi-scale modules. However, over the last 1-2 years, the gain in accuracy has been incremental despite major architectural changes and novel loss function.} Fig. \ref{fig:mae_over_years} compared some well-known crowd counting models showing the mean absolute error over a benchmark dataset and the size of the model parameters. Furthermore, a summary of model architectures, the learning (loss) functions, evaluation metrics, and the training methods (supervised/weakly supervised, etc.) is presented in Table \ref{tab-counting}.


\input{tables/tab_counting2}

\subsection{Challenges and Open Problems}
The accuracy performance (measured using mean absolute error (MAE) metric) of crowd counting models has significantly improved over time on benchmark datasets. For instance, the MAE over ShanghaiTech Part A dataset for MCNN \cite{MCNN_CVPR2016} was 110.2 which is reduced to 66.1 by TransCrowd \cite{TransCrowd_2022}. A similar improvement has been achieved on other benchmarking datasets \cite{khan2022revisiting} as well. Despite the significant improvements in counting accuracy, there still remain several challenges.
\par
\begin{itemize}
\item The problem of scale variations caused due to perspective effects have been overcome by multi-scale architectures, other issues such as occlusions and complex backgrounds are still a major challenge in many complex scenes. \khan{Although some methods propose background segmentation as a preprocessing step, that can increase the complexity of the task.}

\item Second, there is still more room for further improvement in counting accuracy on new benchmark datasets such as UCF-QNRF, JHUCrowd++, and NWPU Crowd datasets due to more challenging scenes e.g., extremely dense crowds, extreme weather and low light conditions, camera blur effects, etc. These conditions are common in the real world and the results typically achieved over previous datasets would not be attainable on these new datasets and thus in the real world. 

\item Third, the commonly used metric in almost all studies is MAE provides an average performance of crowd-counting models. In practice, a model may produce less accurate predictions on difficult examples (e.g., extremely dense and blurry images), but the MAE (being an average over the entire test set) is compensated to be low due to better predictions on easy examples posing a low value of MAE (e.g., less dense and clear images).

\item Fourth, as the benchmark datasets become larger and more challenging over time, the resulting models to achieve better performance over these benchmarks become deeper. This means more complexity to run these models in real-time, especially on resource-limited edge devices. This will create potential bottlenecks in edge-based crowd-monitoring solutions. Lightweight models are being developed \cite{Khan2022DroneNet, Khan2023LCDnet}, but these provide limited accuracy in very dense crowds.
\end{itemize}

\khan{We urge the requirement of more scenario-specific datasets to build reliable models for production. Models trained on generic datasets typically require a long training time and yet lack generalization capabilities.}

\section{Object Detection and Tracking}

There are many situations in which one is interested to detect an object of interest in a video frame and then track it over time in consecutive frames. Detecting a particular object class (e.g., pedestrians, vehicles, men, women, etc.) in a single image or video frame is called object detection, whereas identifying an individual object or a set of objects in consecutive video frames from a single camera or frames from multiple cameras is called object tracking.

\subsection{Methods and State-of-the-art}
Object detection is a well-researched problem that gained significant attention in computer vision and several prominent models are developed over time that achieved state-of-the-art performance. These models are typically divided into two categories i.e., anchor-based methods and anchor-free models. 
Anchor-based methods use a predefined set of anchor boxes placed over the entire image and predict the final set of boxes around the detected objects. These models provide better accuracy and are widely deployed e.g., Faster-RCNN \cite{FasterRCNN_2015}, SSD \cite{SSD_2015}, and YOLO (v3) \cite{YOLOv3_2018}. \rev{Anchor-based methods adapt well to various scales and aspect ratios and work well in complex scenes. However, their performance is greatly affected by various factors such as sizes, aspect ratios, the number of anchor boxes, and shape variations \cite{FCOS_2019}. The predefined anchor boxes require manual design and careful calibration, which can be time-consuming and computationally expensive. However, anchor-based methods have been widely adopted. }

Anchor-free methods are relatively new and more efficient than anchor-based methods. They use the keypoint detection approach. For example in CornerNet \cite{CornerNet_2018}, the model predicts the top-left and bottom-right corners around objects to draw the final bounding box. In CenterNet \cite{CenterNet_2019}, a single point at the center of the object is detected to draw the bounding box.
\par
A fully convolutional one-stage (FCOS) detector is another single-stage anchor-free method proposed in \cite{FCOS_2019} that detects objects using per-pixel prediction that achieve comparable performance to anchor-based methods (Faster-RCNN, YOLOv2, etc.) and outperforms previous anchor-free methods (CornerNet, etc.).
\rev{Anchor-free methods are simpler in design and implementation, and computationally efficient. However, these models perform poorly in accurately localizing small objects or objects with complex shapes, especially in dense and overlapping instances. They also require more training data to sufficiently train.}

\begin{figure}
    \centering
    \includegraphics[width=0.6\textwidth]{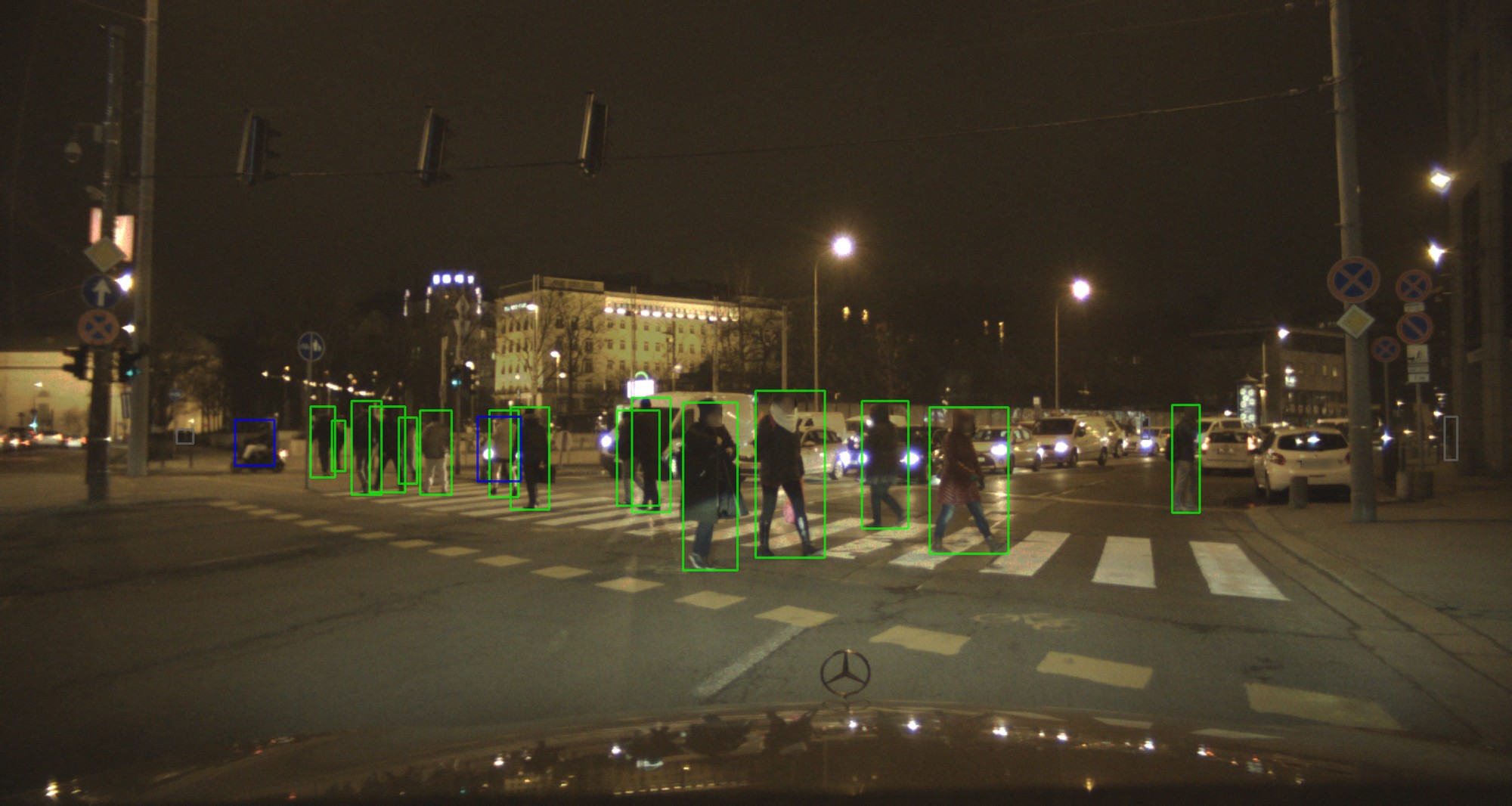}
    \caption{Object detection sample from EuroCity Persons dataset \cite{eurocity_dataset}.}
    \label{fig:obj_detection}
\end{figure}

More recently, non-CNN methods based are getting attention in detection tasks. The vision transformer (ViT) model is proposed in \cite{ViT_2022} as an alternative to CNN-based models. The original ViT and other Transformer-based models have shown comparative performance in many tasks compared to several CNN-based models. \rev{However, CNNs are still considered as defacto-methods for detection tasks due to their fast learning capabilities (training and fine-tuning) as compared to the transformer models.}
Fig. \ref{fig:obj_detection} depicts a sample output of an object detection model showing bounding boxes around the detected objects (i.e., persons). Table \ref{tab-detection} shows a list of objection detection models in the three categories. 

\rev{
There have been several publicly available datasets for object detection tasks however, the most popular datasets used for benchmarking are Pascal VOC \cite{pascal_voc2012} and MS-COCO \cite{ms_coco}. All mainstream object detection models (e.g., YOLO family, FCOS, CornerNet, etc.) are evaluated over these datasets, making the fair benchmarking of model accuracy.
}

\begin{figure}
    \centering
    \includegraphics[width=0.7\textwidth]{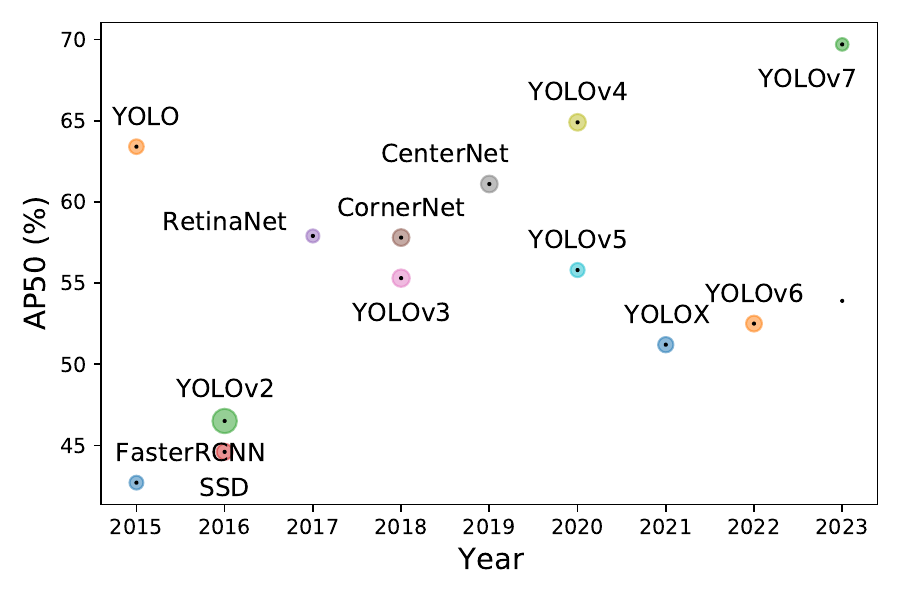}
    \caption{Average precision (AP50) of well-known object detection models on MS COCO dataset (except YOLO which is evaluated on Pascal VOC dataset).} \label{fig:ap_over_years}
\end{figure}

\input{tables/tab_counting2}
Fig. \ref{fig:ap_over_years} shows a performance comparison of various object detection models discussed in this section.

\subsection{Challenges and Open Problems}
Both anchor-based and anchor-free methods have achieved significant performance gains (in production) in the past few years, but both methods still face some intrinsic limitations in the context of crowd scenes. While some of these challenges are generic to any computer vision tasks, some are more severe in crowded scenes given as follows:
\begin{itemize}
\item Viewpoint variation -  when an object looks different when captured from different angles.
\item Object deformation - when an object appears in different shapes in the same frame or in consequent frames of a video (e.g., a person bending down).
\item Severe occlusions - when one or more objects are partially not visible in an image due to overlapping with other objects in front of them.
\item Illuminations - when there are large variations of brightness values of pixels in images.
\item Clutters - when an image contains many or large objects other than objects of interest.
\end{itemize}
The aforementioned issues are very likely to be encountered in many real-world scenarios. 
\khan{Although augmentation techniques play a role to improve the models' performance in learning such challenging environments, there is still no standard method that solves these problems in all scenarios. It is worthy noting that the existing models for object detection are not well-suited for crowd environments and hence despite finetuning produce poor results. This is a serious concern that hinders the adoption of serious surveillance applications (e.g., that can be used by law enforcement agencies).
}

\section{Motion Analysis}

Understanding crowd dynamics can provide more useful information in addition to the crowd count and density estimates. For instance, one may be interested to know whether the crowd is stationary or moving. For a moving crowd, it will be interesting to understand the crowd flux and other patterns related to the crowd movement including trajectory, direction, velocity, etc. It also includes detecting stationary groups in the crowd. Motion analysis has many interesting applications e.g., access control, human identification, congestion analysis, and multi-camera interactive surveillance \cite{hu_2004}. Motion analysis may refer to recognizing the movement of body parts of a person (e.g., gestures, actions, etc.), but in the context of the crowd, it is often referred to as the coherent motion of a group of individuals. Crowd motion analysis is of great interest in understanding crowd behavior analysis and scene understanding e.g., categorizing a crowd as a stationary or moving crowd, crowd trajectory predictions, crowd flux analysis, and crowd motion patterns analysis. Motion analysis may also include studying abnormal motion behavior \cite{Gupta_2019}.

\subsection{Methods and State-of-the-art}
\begin{figure}
    \centering
    \includegraphics[width=0.8\textwidth]{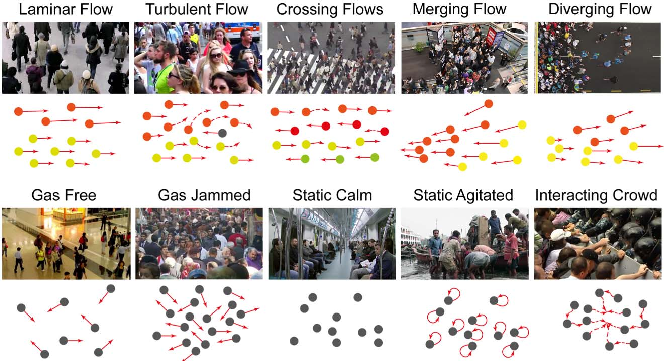}
    \caption{Crowd motion patterns in Crowd 11 dataset \cite{Crowd11_dataset}.}
    \label{fig:motion}
\end{figure}

\rev{Crowd motion analysis methods aim to detect, track, and analyze motion patterns (e.g., in Fig. \ref{fig:motion}) to infer important insights about the dynamic behavior of a crowd.}
Technically, motion analysis includes tempo-spatial analysis of the crowd. There have been several manual and end-to-end automated methods and mathematical models for crowd movement statistics. 
\rev{
Traditionally, motion analysis used methods such as motion segmentation (pixel-wise separation of moving objects from the background), temporal differencing (pixel-wise differences consecutive frames), and optical flow (using flow vectors of moving objects over time) \cite{Hu2004ASO}.
}
For instance, a running count of people's trajectories passing through user-defined lines in a scene is used to measure crowd flows. Several flows can be further integrated over multiple spatial and temporal windows.
Crowd motion is represented in different ways e.g., optical flow (movement of each pixel from one frame to another), particle flow (moving grid of particles with optical flow through numerical integration), streaklines flows (traces left as line upon injection of colored material in the flow.), spatiotemporal features, tracklets (a fragment of trajectory obtains by a tracker with a short period of time), etc \cite{Saqib2019}.

\rev{
Motion segmentation-based methods can be easily implemented using background subtraction or may use methods such as clustering or graph-based approaches. These methods are more sensitive to noise and variations in lighting conditions and poorly perform in the presence of occlusions and recognizing complex motion patterns.
Optical flow-based methods capture the motion of every pixel. The algorithms are computationally efficient allowing real-time performance. However, these methods suffer from inaccuracies to capture small motions between frames, frames with repetitive patterns, and significant scale variation.
}
Authors in \cite{Rabaud_2006} focus on counting moving objects in a crowd by detecting independent motions. Similarly \cite{Lin_2009, Gupta_2019} studies the global motion patterns in crowds. \cite{Ali_2008} studies the segmentation of crowd flows whereas \cite{Hu_2008} uses optical flows to learn the crowd motion patterns.

\subsection{Challenges and Open Problems}
\begin{itemize}
\item Most research on crowd motion analysis is based on the representation of pictorial information such as color, texture, etc. However, these methods only provide some very basic analysis of motion patterns and do not sufficiently provide semantic information on crowd motion.
\khan{Automated motion analysis not only requires large-scale video datasets but also efficient methods to extract both micro and macro statistics of crowd motion. The existing methods seriously lack the sophistication required for real-world implementations.}
\item To extract more high-level and intuitive motion information, models need to learn automatic semantic features and relationships between low-level pictorial features and high-level semantic features.
\end{itemize}

\section{Behavior, Activity and Context Recognition}

\begin{figure}
    \centering
    \includegraphics[width=0.8\textwidth]{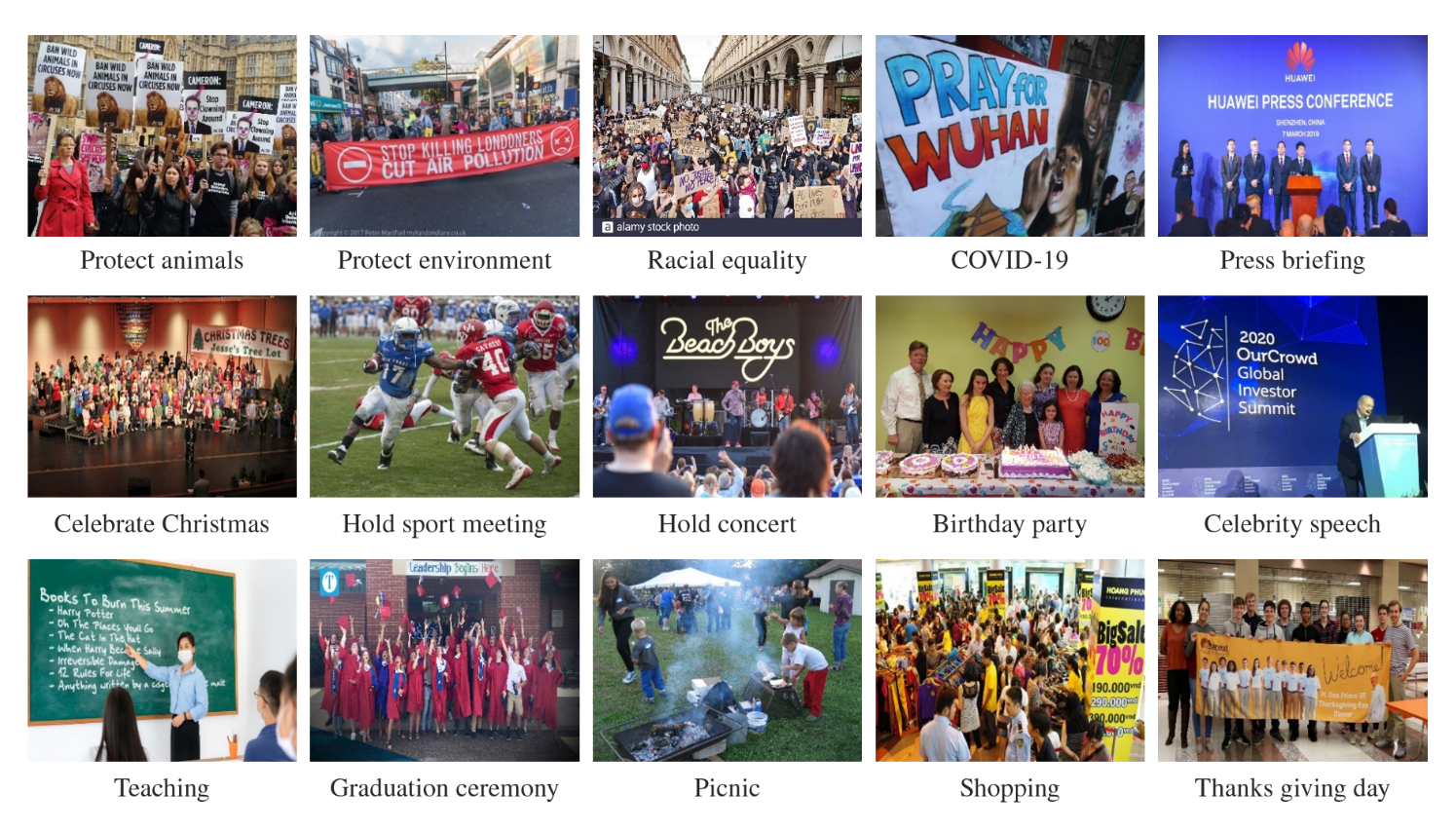}
    \caption{Activity recognition with several crowd activities \cite{Wang2022_KnowledgeMining}.}
    \label{fig:activity}
\end{figure}

Behavior analysis refers to the analysis of the crowd as a whole or a portion of the crowd on a longer time scale (i.e., minutes to hours). It may involve simpler tasks such as identifying the state of the crowd's behavior e.g., calm, active, violent crowd, or may involve complex activity recognition. Activity recognition refers to detecting various grouped activities of crowd members such as protesting, dancing, fighting, etc. usually on a shorter time scale (i.e., seconds to minutes). It can also refer to actions of an individual object e.g., pose estimation. Some examples of group activities are depicted in Fig. \ref{fig:activity}.

The individuals in a crowd may interact with each other and engage with each other in different activities. While activity recognition detects crowd activities using shape, pose, or motion features, context analysis studies the social interaction among group members using time, location, and other contextual information and their relation to the crowd. Context analysis of crowd activity is a more complex problem than activity recognition due to many other factors e.g., environment and other factors related to psychology and sociology.


\subsection{Methods and State-of-the-art}
Activity recognition involves a framework for defining individual and grouped activities and then assigning unique descriptors to each activity. The next step is then to accurately learn the spatial and temporal profiles of each activity either using traditional methods such as mathematical models or machine learning or to apply end-to-end learning using deep learning models.
\rev{Behavior analysis is typically considered as a macroscopic crowd analysis, whereas activity recognition may refer to as microscopic crowd analysis.}
\rev{
In \cite{kok2016crowd}, the authors present common attributes of a crowd (i.e., decentralized, collective motion, emergency behavior) and map these attributes to biological and physical entities.
}

The research on contextual analysis of crowd activities analysis is still very limited due to the inherited complexity of the problem. \rev{Authors in \cite{Benetka2019} present a qualitative analysis of the context in human activity recognition using several attributes such as time and location to establish a spatiotemporal context in the human activity prediction system.}
\rev{
Several works propose the use of low-level feature-based methods such as optical flow.}
In \cite{Khai_2015}, authors studied various aspects of the crowd context analysis. First, discovering meaningful groups in a crowd is modeled as a dominant set clustering algorithm. Second, it uses group context activity (GCA) descriptors of a target person and its semantic neighbors and applies conditional random field  (CRF) and support vector machines (SVM). The authors use two datasets (i) the Collective Activity dataset that involves simple activities i.e., crossing, waiting, queuing, walking, and talking (with two additional activities i.e., dancing and jogging), and (ii) UCLA Courtyard dataset having 10 human activities (Riding, Skateboard, Riding Bike, Riding Scooter, Driving Car, Walking, Talking, Waiting, Reading, Eating, and Sitting).

\subsection{Challenges and Open Problems}
\begin{itemize}
\item Activity recognition and behavior analysis are generally more complex due to the semantic relationships between the detected activities to the human habits. A single behavior can be generally mapped to multiple semantic concepts. Thus, to infer a meaningful semantic behavior, an accurate relationship between the low-level features and the semantic behavior must be established apriori.

\item \rev{It is generally very hard to detect and track individual persons in different crowded scenes \cite{Shao2017}.}

\item Some key problems in behavior analysis include applying background knowledge and reasoning theory to correctly define natural language descriptors to semantic behaviors and then learning these behaviors from the transformations of the object in a scene at different levels. 

\item It is generally desired to use multiple cameras in crowd surveillance. However, it is also more challenging to apply data fusion from multiple sources (e.g., cameras or other sensors) which involves automatically inferring human activities and behaviors from multiple features (rather than images or frames). \khan{Feature extraction and fusion from multiple camera sources also require significant hardware resources and large-scale adoption of such systems can be slower.}

\item Context recognition in crowd analysis is of major importance in a fully automatic crowd surveillance application. It mostly involves multi-modal data from several sources including camera outputs (images or videos) as well as data from sources such as social media (tweets, Facebook posts, comments, etc.), and real-time check-in/checkout records from venues such as airports, metro stations, event venues, etc. The rich information obtained from these diverse sources can provide more accurate contextual information on crowd activities and more accurate semantic descriptors. However, the research in this area is still very limited, and existing works only touch the surface of this broad and deep area in crowd analysis.
\end{itemize}
 
\vspace{-2em}

\section{Anomaly Detection}

\begin{figure}
    \centering
    \includegraphics[width=0.6\textwidth]{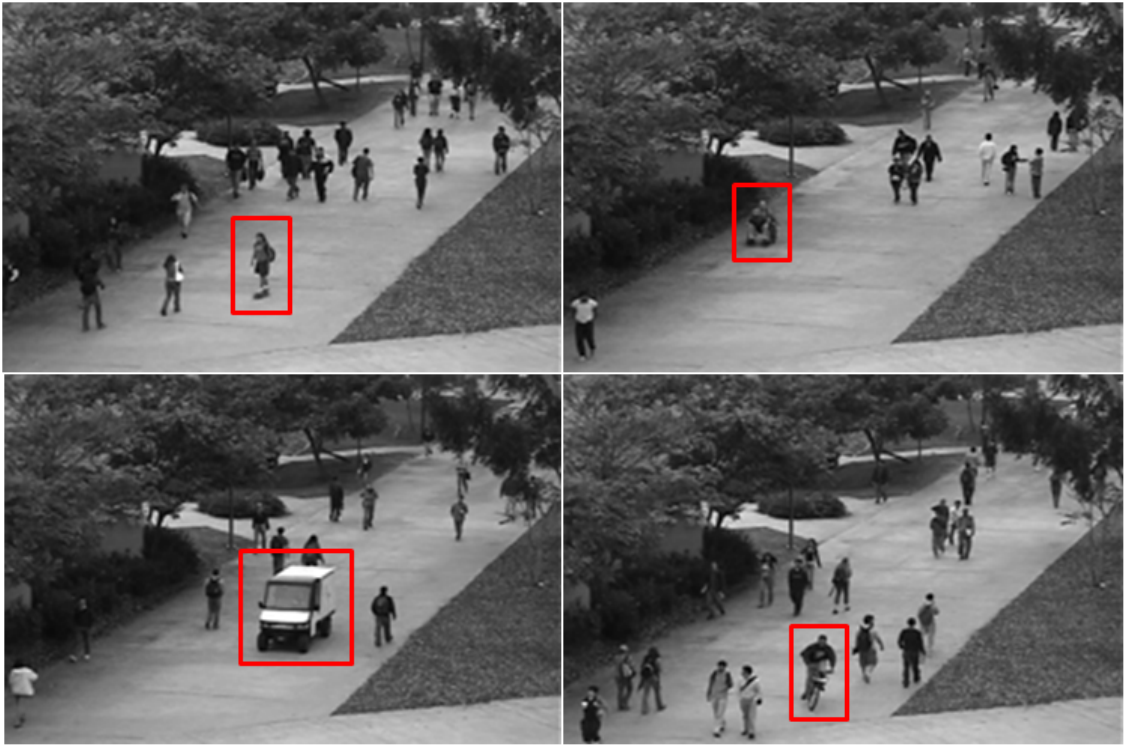}
    \caption{Anomaly examples in pedestrian spaces e.g., wheelchair, skater, biker, and cart. \cite{anomaly_dataset}.}
    \label{fig:anomaly}
\end{figure}

Anomaly detection refers to finding anomalous (aka abnormal) events and has many significant applications e.g., crime detection, traffic violations, abandoned objects detection, weapons detection, etc. In the context of crowd monitoring, anomalies are found using spatiotemporal feature analysis of the video frames as well as in a single image. Fig. \ref{fig:anomaly} shows example anomalies in an outdoor scene.

\subsection{Methods and State-of-the-art}
Crowd anomaly detection is a challenging task mainly due to the rare occurrence of such events and thus lack of sufficient data on anomalous events. \rev{The definition of anomaly is subjective and the same event can be classified as normal and abnormal at different times and places.}
Anomalies can be segregated into (i) point anomaly (when a single entity or person looks or behaves differently than other entities in the scene), (ii) contextual anomaly (when an entity or object is treated as abnormal in a specific contextual situation or environment), and (iii) collective anomalies (when a group of instances behaves abnormally than the rest of the entities in the scene). \rev{Anomaly detection is also classified as a global anomaly (does the scene/frame has anomaly) versus local anomaly (the localization of the anomaly in the frame or video).}
\par

The most basic and commonly used approach to detect anomalies is to train a one-class classifier (OCC) trained with data containing normal examples. The OCC model trained with a sufficiently large number of normal training instances can predict abnormal events. \rev{However, even collecting data with all normal events is not easy.}

\rev{
Anomaly detection in crowded scenes has been addressed in numerous ways using a variety of algorithms including classical machine learning schemes e.g., k-means and SVMs \cite{Yang2019DeepLA}, GMM, etc., as well as deep learning methods e.g., CNNs \cite{Joshi2021ACB, Pang2020SelfTrainedDO, Wu2020VideoAD}, LSTM \cite{Esan2020AnomalousDS}, GANs \cite{Luo2017ARO, Chen2021NMGANNG}, and autoencoders (AEs) \cite{Simonyan2014, Pawar2021ApplicationOD}, bag-of-words (BOW) method, and physics-inspired approaches \cite{Wu2010}, etc. These methods are often inter-related and the exact taxonomy is difficult. In BOW method, spatiotemporal visual features are extracted to detect abnormal events \cite{Javan2013}. BOW method for anomaly detection may not capture complex anomalies and can be used to train on specific anomalies.
Datasets for anomaly detection include UCSD \cite{UCSD_anomaly}, PETS2009, UMN \cite{UMN_anomaly}, CUHK-Avenue \cite{CUHK_Avenue_dataset}, and ShanghatTech Campus \cite{ShanghaiTech_Campus_dataset}.
}

\subsection{Challenges and Open Problems}
Anomaly detection is very challenging due to several reasons including data availability,  compute power requirement, fairness, and generalization. Some of the open problems and challenges are listed as follows.
\begin{itemize}
\item First, there is no universal definition of an abnormal event i.e., an event that is considered abnormal may be considered abnormal in a different context. For example, a person carrying a gun is abnormal but becomes normal when the person carrying the gun is a police constable. Thus, the context is always significant in anomaly detection which makes the anomaly detection problem very challenging. \khan{This is a serious challenge that must be tackled first to enhance the outcomes of research on anomaly detection. Contribution from government and private law enforcement agencies can play significant in data acquisition and annotation. Without sufficient data and standard definitions of crowd anomalies, the research outcomes will be significantly limited.}
\item Second, there is a lack of good datasets for anomaly detection. The existing datasets cover only a small number of anomalies. The methods to label data in these datasets also vary (frame-level anomalies, video-level anomalies, segment-level anomalies, etc.).
\item Third, it is not easy to create large datasets with diverse anomalies because several abnormal events can not be obtained beforehand. \rev{Video datasets for anomalies are sparse, and not typically accessible publicly. The publicly available datasets are mostly captured in the same location and capture very few anomalies.}
\item Fourth, most real-world anomalies can be better captured in video sequences, thus the anomaly detection datasets are typically of very huge size (as compared to other computer vision tasks such as object classification or detection). As a result, huge computational resources are required to train deep-learning models over tons of videos.
\end{itemize}

 \vspace{-2em}

\section{Crowd Prediction}
\rev{
Crowd prediction refers to predicting in advance crowd accumulation in a particular region. Predicting crowds ahead of time has key significance in various scenarios and applications such as events, detecting incidents, tourism attractions, etc. Crowd prediction typically is complex and often leverages multi-modal data from various sources, such as surveillance cameras, social media, mobile phone signals, sensors, etc. Machine learning and statistical modeling techniques are employed to learn and make predictions about future crowds. The process may relate to specific crowd events or gatherings, to estimate the duration of the event, forecast peak crowd hours, or anticipate the popularity of certain areas or attractions within the event.
}
A crowd in specified administrative areas can emerge typically in different ways e.g.,
\textbf{Fountainhead:} crowd emerges from a single direction and  spread over different directions just like a fountain,
\textbf{Bottleneck:} crowd emerges from different directions and assembles at a single point, and
\textbf{Lane:} crowd emerges at a uniform rate and moves in a single direction.
Crowd prediction involves forecasting the aggregated flows (incoming and outgoing) in a specified region. Further information such as the origin and destination of crowd flows are also studied as part of the crowd prediction problem.

\subsection{Methods and State-of-the-art}
\rev{Crowd prediction can be simply modeled as a time series forecasting problem to regress an increase in crowd density over time starting from a no-crowd condition or can also use spatial statistics of crowds to predict crowds in spatial regions.} Traditionally, it uses moving averaging methods such as Autoregressive Integrated Moving Average (ARIMA) \cite{Liu2021}. However, these methods fail to capture the complex temporal and spatial dependencies despite several feature engineering techniques. To cope with complex dependencies, deep learning methods including CNN \cite{Song2020}, LSTM \cite{Li2019}, and graph neural networks (GNNs) \cite{Li2022} have been proposed.
\rev{
Although multi-modal data can be used for predicting crowds, a sufficient quantity of such datasets does not exist. The use of LSTM and CNNs together has been a more promising method to predict crowds over shorter time periods. However, over longer time spans, forecasting methods (e.g., ARIMA) are applied.
}

\subsection{Challenges and Open Problems}
\begin{itemize}
\item Crowd prediction can either be logically concluded from the analysis of crowd motion patterns or predicted directly using time series prediction models. \rev{However, the research in this area is very limited, and very few works are found often using deep learning methods such as CNN and LSTM.}
\item Crowd prediction has both spatial and temporal dependencies and generally, it is quite difficult to predict crowd accumulation in a region over long periods.
\item The prediction in fountainhead and bottleneck is more difficult than a lane pattern.
\item \rev{Due to the spatial dependency, crowd flow prediction in irregular-shaped regions can be more challenging than in regular-shaped regions.}
\end{itemize}

Lastly, Table \ref{tab-summary} provides a summary of the aforementioned six crowd analysis areas, some common examples of tasks in each area, and open research challenges.

\input{tables/tab_summary2}

\section{Conclusion and Future Insights} \label{sec:conclusion}
This article defines six major areas of visual crowd analysis which together form a full-fledged automated crowd-monitoring system. Each of these six areas involves a different level of complexity, and thus the state-of-the-art greatly varies in these. \khan{For instance, crowd counting and detection are two areas with significantly improved results achieved recently over large benchmark datasets.} Typically, existing real-world crowd-monitoring implementations cover these tasks. \rev{The benchmarking in counting and detection is also very clear and can be easily compared. However, there are still areas requiring significant attention. In counting and density estimation, the common regression loss function (Euclidean loss or MSE) has been used. However, over time, several other loss functions (e.g., OT loss, AP Loss, PRA Loss, etc.) have been proposed with reportedly improved performance. However, these loss functions could not be continued in the subsequent works and many recent works consistently used the original MSE loss function. It is encouraged to evaluate these loss functions over different datasets and several mounting models to conclude their potential benefits in this domain. We also encourage testing the potential performance gain of using curriculum learning (CL) in crowd-counting tasks with appropriate curriculum strategies.}

\rev{Motion analysis that focuses on crowd-level mobility statistics often uses flow-based and spatiotemporal features. Flow-field models are relatively more studied. However, clustering models are gaining attention due to their performance in more crowded scenes. Motion analysis tasks vary such as detecting crowd source/sink, trajectory finding, speed, etc. The benchmarking in this application domain is less coherent due to the nature of the task and multiple objectives being considered in existing studies.
}
\khan{
Behavior analysis and anomaly detection which are sometimes overlapping are the most complex tasks and the progress in these tasks is still very limited and scattered in terms of methods, approaches, assumptions, and objectives despite their major importance in several use cases.}
\rev{
The lack of definition of anomalies, activities, and behavior causes researchers to use different objectives and evaluation metrics which makes benchmarking unfair. We suggest that future research should focus on developing common definitions of activities and anomalies considering context and enhancing existing datasets as well as creating larger and balanced datasets. We also believe that in many scenarios, local anomalies would be required rather than global anomalies which shall make the task easier to learn; however, the variations in environment and context will make the cross-domain transfer-learning challenging. Physics-inspired approaches (e.g., energy models) are interesting directions to implement anomaly detection.}
\par
We envision major advances in the near future in the under-explored areas due to the recent developments in generative AI which will be helpful to cope with the need for more training data. Furthermore, crowd analysis in real-time typically requires fast inference. In CCTV-based surveillance, it is more convenient to perform all processing and inference tasks on a local server due to the high-speed wired connectivity option however, in aerial surveillance (using drones) on-device processing and inference may be more convenient in some cases. Thus, lightweight crowd analysis models will be preferred.

%% file: tables/tab_surveys2.tex
\begin{table*}[!h]
\centering
\small
\caption{Related studies on crowd visual analysis.}
\label{tab-surveys}
\begin{tabular}{p{3cm} p{2.5cm} p{8cm}} \toprule \\
Ref & Topic  & Brief Scope  \\[0.5em] \midrule
& \\[-0.5em]
\cite{Sindagi_2018} & Crowd counting &Methods, model architectures, datasets, future insights \\[0.5em]
\cite{Naveed_2020} & Crowd counting &Methods, model architectures, performance evaluation. \\[0.5em]
\cite{Gao_2020} & Crowd counting &Model architectures and categorization, datasets, evaluation metrics, challenges, and insights. \\[1.5em]
\cite{Gouiaa_2021} & Crowd counting &Methods, the taxonomy of CNN models, datasets, and applications. \\[1.5em]
\cite{Fan_2022} & Crowd counting &Methods, applications, and models. \\[0.5em]
\cite{khan2022revisiting} & Crowd counting &Datasets, model architecture, loss functions, evaluation metrics, insights \\[1.5em]

\cite{Hu2004ASO} &Motion analysis &Motion analysis, behavior analysis, moving object classification and identification. \\[1.5em]
\cite{Luca2020ASO} &Motion analysis &Flow prediction, next-location prediction, flow generation, and trajectory generation. \\[1.5em]

\cite{Kumar2021CrowdBM} &Behavior analysis &Methodologies for the organized crowd and non-organized crowd, datasets. \\[1.5em]

\cite{Modi2022ASO} & Anomaly detection &Methods for anomaly detection with merits and demerits. \\[0.5em]
\cite{Sharif2022DeepCA} &Anomaly detection & Methods, algorithms, datasets, and metrics with details analysis and comparisons.  \\[1.5em]

\cite{Li2015CrowdedSA} & Crowd analysis &Models, datasets, algorithms, and evaluation of crowd motion analysis, behavior recognition, and anomaly detection.  \\[1.5em]
\cite{HY2017CrowdBA} &Crowd analysis &Density estimation, motion detection and tracking, and behavior recognition. \\[1.5em]
\cite{Zhang2018PhysicsIM} &Crowd analysis &Physics-inspired methods for crowd video analysis. \\[0.5em]
\bottomrule
\end{tabular}
\end{table*}

%% file: tables/tab_counting2.tex
\begin{table*}[!h]
\centering
\caption{A summary of research efforts in crowd counting, categorized in distinct categories i.e., model architecture, loss functions, metrics, and training methods.} 
\label{tab-counting}

\begin{tabular}{c p{10cm}} \toprule \\
Research Area &Summary of the SOTA  \\[0.5em] \midrule
& \\[0em]
Model Architectures
&Single-column \cite{CrowdCNN_CVPR2015}, multi-column \cite{MSCNN_ICIP2017}, Encoder-decoder \cite{TEDnet_CVPR2019}, pyramid, Vision transformers \cite{TransCrowd_2022} \\[1em]

Loss functions 
&Euclidean loss \cite{MCNN_CVPR2016}, Combination loss \cite{CrowdCNN_CVPR2015}, Curriculum Loss \cite{SGANet_IEEEITS2022}, Composite loss \cite{CompositionLoss_2018}, AP Loss \cite{ASNet_CVPR2020}, PRA Loss \cite{ASNet_CVPR2020}, SCL Loss \cite{TEDnet_CVPR2019}, OT \cite{DMCount_NeuroIPS2020}\\[1em]

Evaluation metrics
& MAE, MSE, GAME \cite{GAME_metric2015}, SSIM \cite{CSRNet_CVPR2018}, PSNR \cite{CSRNet_CVPR2018}, PAME \cite{}, PMSE \cite{}, MPAE \cite{} \\[1em]

Training methods
& Supervised/weakly-supervised \cite{Yang_ICCV2020, TransCrowd_2022, MATT_Elsevier_PR2020}, curriculum learning \cite{Khan2023LCDnet} \\[1em]
\bottomrule
\end{tabular}
\end{table*}

%% file: tables/tab_summary2.tex
\begin{table*}
\caption{Crowd analysis tasks and Open research problems.} \label{tab-summary}
\centering
\small
\begin{tabular}{p{2.5cm} p{6cm} p{6cm}} \toprule 
&& \\[-0.5em]
\textbf{Crowd Analysis} & \textbf{Example Tasks}  & \textbf{Open Challenges} \\[1em] \midrule 

Crowd Counting
& Find the total headcount in a frame. \newline 
crowd localization in a large region. \newline
& \rev{Handling severe occlusions, scale variations in highly dense crowd images, perspective distortion, multi-view counting.}
\\ [1em]

Object Detection 
& Detect and localize objects of interest. \newline 
Track objects in a video.
& \rev{Viewpoint variations, Object deformation, severe occlusions, clutter, illumination, etc.}
\\ [1em]

Motion Analysis
& Observe crowd collectiveness. \newline
Is the crowd slow/fast-moving? \newline
Where is the crowd heading? at which speed? \newline
Find trajectories and motion patterns. \newline
& \rev{Complex semantic features, complex relationship between low-level pictorial features with high-level semantic features.}
\\ [1em]

Behavior/Activity Analysis
& Is the crowd calm or active? \newline
Are people protesting, dancing, fighting, etc?
&\rev{Complex mapping of low-level features to activities, semantic relationship and mapping of detected activities to human behavior, standard activity definitions to semantic behavior.}\\ [1em]

Anomaly Detection
& Detect abandoned objects. \newline 
Detect any weapons or crimes? \newline
Detect intrusion.
& \rev{Lacks of a unified definition of anomalies, lack of realistic datasets, the huge size of video datasets \footnote{An extensive list of challenges can be found in \cite{Sharif2022DeepCA}}.}
\\ [1em]

Crowd prediction
& Detect if a crowd is expected in a region? \newline 
How crowd accumulates over time?
& \rev{Difficult to capture dual (spatial and temporal) dependencies, complex motion patterns.} \\ [1em]
\bottomrule
\end{tabular}
\end{table*}